# STATISTICAL FUNCTION TAGGING AND GRAMMATICAL RELATIONS OF MYANMAR SENTENCES

Win Win Thant[1], Tin Myat Htwe[2] and Ni Lar Thein[3]

[1,3]University of Computer Studies, Yangon, Myanmar
winwinthant@gmail.com
nilarthein@gmail.com
[2]Natural Language Processing Laboratory
University of Computer Studies, Yangon, Myanmar
tinmyathtwe@gmail.com

### ABSTRACT

*This paper describes a context free grammar (CFG) based grammatical relations for Myanmar sentences which combine corpus-based function tagging system. Part of the challenge of statistical function tagging for Myanmar sentences comes from the fact that Myanmar has free-phrase-order and a complex morphological system. Function tagging is a pre-processing step to show grammatical relations of Myanmar sentences. In the task of function tagging, which tags the function of Myanmar sentences with correct segmentation, POS (part-of-speech) tagging and chunking information, we use Naive Bayesian theory to disambiguate the possible function tags of a word. We apply context free grammar (CFG) to find out the grammatical relations of the function tags. We also create a functional annotated tagged corpus for Myanmar and propose the grammar rules for Myanmar sentences. Experiments show that our analysis achieves a good result with simple sentences and complex sentences.*

### KEYWORDS

*Grammatical relations, Function tagging, Naive Bayesian theory, Context-free-grammar*

## 1. INTRODUCTION

Myanmar is an agglutinative language with a very productive inflectional system. This means that for any NLP application on Myanmar to be successful, some amount of functional analysis is necessary. Without it, the development of grammatical relations would not be feasible due to the sparse data problem bound to exist in the training data. It is the process of analyzing an input sequence in order to determine its grammatical structure with respect to a given grammar. Grammatical relations operate at word-level with the assumption that input sentences are pre-segmented, POS tagged and chunked.

The natural language processing community is in the strong position of having many available approaches to solving some of its most fundamental problems [1]. We have taken Myanmar language for information processing. Our approach makes use of two components. They are function tagging and grammatical relations. Function tags are useful for any application trying to follow the thread of the text –they find the 'who does what' of each clause, which can be useful to gain information about the situation or to learn more about the behaviour of words in the sentence [2]. The goal of function tagging is to assign syntactic categories like subject, object, time and location to each word in the text document. In case of function tagging, we use Naive Bayesian theory and the functional annotated tagged corpus. Grammatical relations are the process of analyzing an input sequence in order to determine its grammatical structure with respect to a given grammar. The goal of the second one is to produce the relations of the grammatical structures of the sentences in Myanmar text as a parse tree.

Myanmar is SOV language. It is also a variable phrase order language. The free phrase order feature of Myanmar makes statistical function tagging a challenging task. Function tagging is a part of the Myanmar to English machine translation project. If high quality translation is to be achieved, language understanding is a necessity. One problem in Myanmar language processing is the lack of grammatical regularity in the language. This leads to very complex Myanmar grammar in order to obtain satisfactory results, which in term increases the complexity in the grammatical relation process, it is desired that simple grammar is to be used.

In our approach, we take the chunk level phrase with the combination of POS tag and its category which is the output of a fully described morphological analyzer [3][4], which is very important for agglutinative languages like Myanmar. A small corpus annotated manually serves as training data because the large scale Myanmar Corpus is unavailable at present. Since the large-scale annotated corpora, such as Penn Treebank, have been built in English, statistical knowledge extracted from them has been shown to be more and more crucial for natural language disambiguation [5]. As a distinctive language, Myanmar has many characteristics different from English. The use of statistical information efficiently in Myanmar language is still a virgin land waiting to explore.

Naïve Bayesian is chosen for its simplicity and user-friendliness. Naive-Bayesian classifier make strong assumptions about how the data is generated, and use a probabilistic model that reflects these assumptions [6]. They use a collection of labelled training examples to estimate the parameters of the generative model. Classification of new examples is performed with Bayes' rule by selecting the class that is most likely to have generated the example.

## 2. LITERATURE SURVEY

Blaheta and Johnson [7] addressed the task of function tags assignment. They used a statistical algorithm based on a set of features grouped in trees, rather than chains. The advantage was that features can better contribute to overall performance for cases when several features are sparse. When such features are conditioned in a chain model the sparseness of a feature can have a dilution effect of an ulterior (conditioned) one.

Mihai Lintean and Vasile Rus[8] described the use of two machine learning techniques, naive Bayes and decision trees, to address the task of assigning function tags to nodes in a syntactic parse tree. They used a set of features inspired from Blaheta and Johnson [7]. The set of classes they used in their model corresponds to the set of functional tags in Penn Treebank. To generate the training data, they have considered only nodes with functional tags, ignoring nodes unlabeled with such tags. They trained the classifiers on sections 1-21 from Wall Street Journal (WSJ) part of Penn Treebank and used section 23 to evaluate the generated classifiers.

Yong-uk Park and Hyuk-chul Kwon [9] tried to disambiguate for syntactic analysis system by many dependency rules and segmentation. Segmentation is made during parsing. If two adjacent morphemes have no syntactic relations, their syntactic analyzer makes new segment between these two morphemes, and find out all possible partial parse trees of that segmentation and combine them into complete parse trees. Also they used adjacent-rule and adverb subcategorization to disambiguate of syntactic analysis. Their syntactic analyzer system used morphemes for the basic unit of parsing. They made all possible partial parse trees on each segmentation process, and tried to combine them into complete parse trees.

Mark-Jan Nederhof and Giorgio Satta[10] considered the problem of parsing non-recursive context-free grammars, i.e., context-free grammars that generate finite languages and presented two tabular algorithms for these grammars. They presented their parsing algorithm, based on the CYK (Cocke–Younger–Kasami) algorithm and Earley's alogrithm. As parsing CFG (context-free grammar), they have taken a small hand-written grammar of about 100 rules. They have ordered the input grammars by size, according to the number of nonterminals (or the number of nodes in the forest, following the terminology by Langkilde (2000)).

Kyongho Min and William H. Wilson [11] discussed the robustness of four efficient syntactic error-correcting parsing algorithms that are based on chart parsing with a context-free grammar. They implemented four versions of a bottom-up error-correcting chart parser: a basic bottom-up chart parser, and chart parsers employing selectivity, top-down filtering, and a combination of selectivity and a top-down filtering. They detected and corrected syntactic errors using a system component called IFSCP (Ill-Formed Sentence Chart Parser) described by Min & Wilson (1994), together with a spelling correction module. They tested 4 different lengths of sentences (3, 5, 7, and 11) and 5 different error types, with a grammar of 210 context-free rules designed to parse a simple declarative sentence with no conjunctions, passivisation, or relative clauses.

## 3. MYANMAR LANGUAGE

The Myanmar language is the official language and is more than one thousand years old.

### 3.1. Features of Myanmar Language

Unlike English language Myanmar is syntax of relatively free-phrase-order language. This can be easily illustrated with the example "သူသည် စာအုပ်ကို စားပွဲပေါ်တွင် ထားသည်။" (He places the book on the table) as shown in table 1. All are valid sentences [12].

Table 1. Word order in Myanmar language

| Case | Myanmar Sentences | Word order |
|---|---|---|
| Case 1 | သူ စာအုပ်ကို စားပွဲပေါ်တွင် ထားသည်။ | (Subj-Obj-Pla-Verb) |
| Case 2 | သူ စားပွဲပေါ်တွင် စာအုပ်ကို ထားသည်။ | (Subj-Pla-Obj-Verb) |
| Case 3 | စာအုပ်ကို စားပွဲပေါ်တွင် သူ ထားသည်။ | (Obj-Pla-Subj-Verb) |
| Case 4 | စာအုပ်ကို သူ စားပွဲပေါ်တွင် ထားသည်။ | (Obj-Subj-Pla-Verb) |
| Case 5 | စားပွဲပေါ်တွင် သူ စာအုပ်ကို ထားသည်။ | (Pla-Subj-Obj-Verb) |
| Case 6 | စားပွဲပေါ်တွင် စာအုပ်ကို သူ ထားသည်။ | (Pla-Obj-Subj-Verb) |

In all the cases, subject is သူ (He), object is စာအုပ်ကို (the book), place is စားပွဲပေါ်တွင် (on the table) and verb is ထားသည် (places). From the above example, it is clear that phrase order does not determine the functional structure in Myanmar language and permits scrambling. Myanmar language follows Subject-Object-Verb orders in contradiction with English language.

### 3.2. Issues of Myanmar Language

The highly agglutinative language like Myanmar, nouns and verbs get inflected. Many times we need to depend on syntactic function or context to decide upon whether the particular word is a noun or adjective or adverb or post position [12]. This leads to the complexity in Myanmar grammatical relations. A noun may be categorized as common, proper or compound. Similarly, verb may be finite, infinite, gerund or contingent.

A number of issues are affecting the function tagging for Myanmar language.

- Myanmar phrases can be written in any order as long as the verb phrase is at the end of sentence.
  For example:
  မောင်လှသည် - စာအုပ်တစ်အုပ်ကို - မောင်ဘအား - ပေးသည်။
  Mg Hla - a book - to Mg Ba - gives

(or)
စာအုပ်တစ်အုပ်ကို - မောင်ဘအား - မောင်လှက - ပေးသည်။
a book - to Mg Ba - Mg Hla - gives
(Ma Hla gives a book to Mg Ba.)

- The phrase order of Myanmar language is free. The sentence can be constructed by placing emphatic phrases at the beginning of a sentence.
  For example:
  သူသည်- သတင်းစာကို - ဖတ်သည်။(Subj-Obj-Verb)
  He - newspaper – reads
  (or)
  သတင်းစာကို - သူ - ဖတ်သည်။ (Obj-Subj-Verb)
  newspaper - he - reads
  (He reads the newspaper.)
- The subject or object of the sentence can be skipped, and still be a valid sentence.
  For example:
  ရန်ကုန်သို့သွားသည်။ (Go to Yangon)
- Myanmar language makes prominent usage of particles, which are untranslatable words that are suffixed or prefixed to words to indicate level of respect, grammatical tense, or mood.
  For example:
  မောင်မောင် - **များ** - ပထမ - ဆု - ရ - လျှင် - သူ့မိဘများ - က - အံ့သြ - လိမ့်မည်။
  Mg Mg - particle - first - prize - wins - if - his parents - PPM - surprise - will
  (If Mg Mg wins the first prize, his parents will surprise.)
- In Myanmar language, an adjective can specialize before or after a noun unlike other languages.
  For example:
  သူသည် - **ချမ်းသာသော** - လူ -တစ်ယောက် -ဖြစ်သည်။
  He - rich - man - a - is
  (or)
  သူသည် - လူ - **ချမ်းသာ** - တစ်ယောက် -ဖြစ်သည်။
  He - man - rich - a - is
  (He is a rich man.)
- The subject /object can be another sentence, which does not contain subject or object.
  For example:
  ကလေးများသစ်ပင်အောက်တွင်ကစားနေသည် ကို ကျွန်တော်မြင်သည်။
  (I see the children playing under the tree.)
- The postpositions of subject phrases or object phrases can be hidden.
  For example:
  သူ**သည်**- ဆရာဝန် -တစ်ယောက် - ဖြစ်သည်။
  He - doctor - a - is
  (or)
  သူ - ဆရာဝန် - တစ်ယောက် - ဖြစ်သည်။
  He - doctor - a - is
  (He is a doctor.)
- The postpositions of time phrases or place phrases can be omitted.
  For example:
  သူမ - ကျောင်း - **သို့** - သွားသည်။
  She - school - to - goes
  (or)
  သူမ - ကျောင်း - သွားသည်။

She - school - goes
(She goes to school.)

- The verb phrase can be hidden in a Myanmar sentence.
  For example:
  သူ - မောင်လှ -ပါ။
  He - Mg Hla - particle
  (He is Mg Hla.)

These issues will cause a lot of problem during function tagging, and a lot of possible tags will be resulted.

## 3.3. Grammar of Myanmar Language

Grammar studies the rules behind languages. The aspect of grammar that does not concern meaning directly is called *syntax*. Myanmar (syntax: SOV), because of its use of postposition (wi.Bat), would probably be defined as a “postpositional language”, whereas English (syntax: SVO) because of its use of preposition would probably be defined as a “prepositional language”.

There are really only two parts of speech in Myanmar, the noun and the verb, instead of the usually accepted eight parts (Pe Maung Tin 1956:195). Most Myanmar linguists [13] accepted there are eight parts of speech in Myanmar. Myanmar nouns and verbs need the help of suffixes or particles to show grammatical relations.

For example:
ကျောင်းသူများ**သာ** ဂုဏ်ထူးရသည်။
သူတို့သည် အတန်းထဲမှာ ရှိ**ကြ**၏။

Myanmar is a highly verb-prominent language and that suppression of the subject and omission of personal pronouns in connected text result in a reduced role of nominals. This observation misses the critical role of postposition particles marking sentential arguments and also of the verb itself being so marked. The key to the view of Myanmar being structures by nominals is found in the role of the particles. Some particles modify the word's part of speech. Among the most prominent of these is the particle အ, which is prefixed to verbs and adjectives to form nouns or adverbs.There is a wide variety of particles in Myanmar [14].

For example:
သူတို့သည် မန္တ လေးတွင် ၈ ရက် **တိတိ** လည်ခဲ့သည်။

Stewart remarked that “The Grammar of Burmese is almost entirely a matter of the correct use of particles”(Stewart 1956: xi). How one understands the role of the particles is probably a matter of one's purpose.

## 3.4. Syntacic Structure of Myanmar Language

It is known that many postpositions can be used in a Myanmar sentence. If the words can be misplaced in a sentence, the sentence can be abnormal. There are two kinds of sentence as a sentence construction. They are simple sentence (SS) and complex sentence (CS). In simple sentence, other phrases such as object, time, and place can be added between subject and verb. There are two kinds of clause in a complex sentence called independent clause(IC) and dependent clause (DC).There must be at least one independent clause in a sentence. But there can be more than one dependent clause in it. IC contains sentence’s final particle (sfp) at the end of a sentence [15].

SS=IC+sfp
CS=DC...+IC+sfp

IC may be noun phrase or verb or combination of both.

IC=N... (မျက်မှန်နှင့်ကျောင်းသား)
IC=V (စား)
IC=N...+V (ဘုရားမှာပန်းနဲ့ဆီမီးလှူ)

DC is the same as IC but it must contain a clause marker (cm) in the end.

DC=N...+cm (ကျောင်းကဆရာ+ပဲ)
DC=V+cm (ရောက်+ရင်)
DC=N...+V+cm (စိတ်ထား+ဖြူ+မှ)

## 4. PROPOSED SYSTEM

The procedure of the proposed system is described in the following.

Step1. Accept input Myanmar sentence with segmentation, POS tagging and chunking
Step2. Extract one POS tag and its category from each chunk
Step3. Choose the possible function tags for each POS tag by using Naive Bayesian theory
Step4. Display the sentence with function tags
Step5. Parse the function tags by using CFG rules with the proposed grammar
Step6. Display the parse tree as an output

## 5. CORPUS CREATION

We collected several types of Myanmar texts to construct a corpus. Our corpus is to be built manually. We extended the POS tagged corpus that is proposed in [3]. The chunk and function tags are manually added to the POS tagged corpus. The number of sentences is about 3000 sentences with average word length 15 and it is not a balanced corpus that is a bit biased on Myanmar textbooks of middle school. The corpus size is bigger and bigger because the tested sentences are automatically added to the corpus. In table 2, Myanmar grammar books and websites are text collections. Example corpus sentence is shown in figure 2.

Table 2. Corpus Statistics

| **Text types** | **# of sentences** |
|---|---|
| Myanmar textbooks of middle school | 1200 |
| Myanmar Grammar books | 600 |
| Myanmar websites | 900 |
| Others | 300 |
| Total | 3000 |

VC@Active[မိုးရွာ/verb.common]#CC@CCS[လျှင်/cc.sent]#NC@Subj[ကလေး/n.person,များ/part.number]#NC@PPla[လမ်း/n.location]#PPC@PlaP[ပေါ်တွင်/ppm.place]#NC@Obj[ဘောလုံး/n.objects]#VC@Active [ကန်ကြ/verb.common]#SFC@Null[သည်/sf]။

Figure 2. A sentence in the corpus

## 6. FUNCTION TAGSET

Function tagging is a process of assigning syntactic categories like subject, object, time and location to each word in the text document. These are conceptually appealing by encoding an event in the format of “who did what to whom, where, when”, which provides useful semantic information of the sentences. We use the function tags that is proposed in [16] because it is easier to maintain and can add new language features. The function tagsets are shown in table 3.

Table 3. Function Tagsets

| **Tag** | **Description** | **Example** |
|---|---|---|
| Active | Verb | စားသည် |
| Subj | Subject | သူ |
| PSubj | Subject | သူ |
| SubjP | Postposition of Subject | သည် |
| Obj | Object | ကော်ဖီ |
| PObj | Object | ကော်ဖီ |
| ObjP | Postposition of Object | ကို |
| PIobj | Indirect Object | မလှ |
| IobjP | Postposition of Indirect Object | အား |
| Pla | Place | ရန်ကုန် |
| PPla | Place | ရန်ကုန် |
| PlaP | Postposition of Place | သို့ |
| Tim | Time | မနက် |
| PTim | Time | မနက် |
| TimP | Postposition of Time | တွင် |
| PExt | Extract | ကျောင်းသားများ |
| ExtP | Postposition of Extract | အနက် |
| PSim | Similie | မင်းသမီး |
| SimP | Postposition of Similie | ကဲ့သို့ |
| PCom | Compare | သူ့ဦးလေး |
| ComP | Postposition of Compare | နှင့်အတူ |
| POwn | Own | သူ |
| OwnP | Postposition of Own | ၏ |
| Ada | Adjective | လှ |
| PcomplS | Subject Complement | သူသည်**ဆရာ**ဖြစ်သည် |
| PcomplP | Object Complement | ရွှေကို**လက်စွပ်**လုပ် သည် |
| PPcomplO | Object Complement | ထွန်းထွန်း |
| PcomplOP | Postposition of Object Complement | ဟု |
| PUse | Use | တုတ် |
| UseP | Postposition of Use | ဖြင့် |
| PCau | Cause | မိုး |
| CauP | Postposition of Cause | ကြောင့် |
| PAim | Aim | အမေ့ |
| AimP | Postposition of Aim | အတွက် |
| CCS | Join the sentences | လျှင် |
| CCM | Join the meanings | ထို့ကြောင့် |
| CCC | Join the words | နှင့် |
| CCP | Join with particles | ကို |
| CCA | Join as an adjective | မည့် |

## 7. PROPOSED GRAMMAR FOR MYANMAR SENTENCES

Since it is impossible to cover all types of sentences in Myanmar language, we have taken some portion of the sentence and try to make grammar for them. Myanmar is free-phrase-order language. In Myanmar language, we see that one sentence can be written in different forms for the same meaning, i.e. the positions of the tags are not fixed. So we cannot restrict the grammar rule for one sentence. The grammar rule may be very long, but we have to accept it. The grammar rule we have tried to make, may not work for all the sentences in Myanmar language because we have not considered all types of sentences. Some of the sentences are shown below, which are used to make the grammar rules.

| | |
|---|---|
| သူ-သည်-ကျောင်း-သို့-သွား-သည်။ | (Subj-Pla-Verb) |
| သူ-သည်-ကျောင်းသားတစ်ယောက်-ဖြစ်-သည်။ | (Subj-PcomplS-Verb) |
| ကောင်စီဝင်-အဖြစ်-သူ့-ကို-လူထု-က-ရွေး-သည်။ | (PcomplO-Obj-Subj-Verb) |
| မောင်လှ-သည်-ခွေး-ကို-တုတ်-ဖြင့်-ရိုက်-သည်။ | (Subj-Obj-Use-Verb) |
| သူ-သည်-ဆရာ့-ကို-စာအုပ်-ပေး-သည်။ | (Subj-Obj-Iobj-Verb) |
| သူမ-သည်-လူနာများ-ကို-ဆွေမျိုးများ-ကဲ့သို့-ပြုစု-သည်။ | (Subj-Obj-Sim-Verb) |
| ကလေးများ-သည်-အဖော်-ကြောင့်-ပျက်စီး-သည်။ | (Subj-Cau-Verb) |
| သစ်ရွက်တို့-သည်-တပေါင်းလ-၌-ကြွေ-သည်။ | (Subj-Tim-Verb) |
| တရားသူကြီး-သည်-ခိုးမှု-ကို-တရားရုံး-၌-နံနက်-က-စစ်ဆေး-သည်။ | (Subj-Obj-Pla-Tim-Verb) |
| အမေသည်-သူ့သားအတွက်-မုန့်ကို-ဈေးမှ-မနက်က-ဝယ်ခဲ့သည်။ | (Subj-Aim-Obj-Pla-Tim-Verb) |

Our proposed grammar for Myanmar Sentences:

| | |
|---|---|
| Sentence | →I-sent \| I-sent CC I-sent \| Obj-sent I-sent \| Subj-sent I-sent |
| I-sent | →Subj Obj Pla Verb \| Subj Verb \| Com Pla Verb |
| CC | →CCA \| CCS \| CCM |
| Subj -sent | →I-sent CCA Subj |
| Obj -sent | →I-sent CCA Obj |
| Subj | →PSubj SubjP |
| Subj | →Subj |
| Obj | →PObj ObjP |
| Obj | →Obj |
| Pla | →PPla PlaP |
| PcomplO | →PPcomplO PcomplOP |
| Use | →PUse UseP |
| Sim | →PSim SimP |

## 8. NAIVE BAYESIAN CLASSSIFIER

Before one can build naive Bayesian based classifier, one needs to collect training data. The training data is a set of problem instances. Each instance consists of values for each of the defined features of the underlying model and the corresponding class, i.e. function tag in our case. The development of a naive Bayesian classifier involves learning how much each function tag should be trusted for the decisions it makes [17]. In probability estimation for Naive Bayesian classifiers, namely that the attribute values are conditionally independent when the target value is given. Naive Bayesian classifiers are well-matched to the function tagging problem.

The Naïve Bayesian classifier is a term in Bayesian statistics dealing with a simple probabilistic classifier based on applying Bayes' theorem with strong (naïve) independence assumptions. It assumes independence among input features. Therefore, given an input vector, its target class can be found by choosing the one with the highest posterior probability.

## 8.1. Function Tagging by Using Naïve Bayes Theory

The labels such as subject, object, time, etc. are named as function tags. By function, it is meant that action or state which a sentence describes. The system operates at word-level with the assumption that input sentences are pre-segmented, pos-tagged and chunked.

Each proposed function tag is regarded as a class and the task is to find what class/tag a given word in a sentence belongs to a set of predefined classes/tags. A feature is a POS tag word with category. The category of a word is added to the POS tag to obtain more accurate lexical information. It can be formed from the features of that word. For example, noun has 16 categories such as animals, person, objects, food, location, etc. There are 47 categories in our corpus. We show some features of Myanmar words as shown in table 4.

Table 4. Features

| **Feature** | **English** | **Myanmar** |
|---|---|---|
| n.food | apple | ပန်းသီး |
| pron.possessive | his | သူ့ |
| ppm.time | at | တွင် |
| adj.dem | happy | ပျော်ရွှင်သော |
| part.support | can | နိုင် |
| cc.mean | so | ထို့ကြောင့် |
| v.common | go | သွား |
| sf.declarative | null | ၏ |

In Myanmar language, some words have same meaning but in different features as shown in table 5.

Table 5. Same word with different features

| **Feature** | **English** | **Myanmar** |
|---|---|---|
| cc.chunk | and | နှင့် |
| ppm.compare | with | နှင့် |
| ppm.use | with | နှင့် |

A class is a one of the proposed function tags. Same word may have different function tags as shown in table 6.

Table 6. Function tags

| **Function tags** | **English** | **Myanmar** |
|---|---|---|
| PcomplS | He has a **house**. | အိမ် |
| PPla | He lives in a **house**. | အိမ် |
| PSubj | A **house** is near the school. | အိမ် |
| PObj | He buys a **house**. | အိမ် |

There are many chunks in a sentence such as NC (noun chunk), PPC (postpositional chunk), AC (adjectival chunk), RC (adverbial chunk), CC (conjunctional chunk), SFC (sentence's final chunk) and VC (verb chunk). The chunk types are shown in table 7.

Table 7. Chunk types

| No. | Chunk Type | Example |
|---|---|---|
| 1 | Noun Chunk | NC[သူတို့/pron.person] |
| 2 | Postpositional Chunk | PPC[သည်/ppm.subj] |
| 3 | Adjectival Chunk | AC[ရဲရင့်/adj.dem] |
| 4 | Adverbial Chunk | RC[လျင်မြန်စွာ/adv.manner] |
| 5 | Conjunctional Chunk | CC[သို့မဟုတ်/cc.chunk] |
| 6 | Sentence Final Chunk | SFC[၏/sf.declarative] |
| 7 | Verb Chunk | VC[ကူညီ/v.common] |

A chunk contains a Myanmar head word and its modifier. It can contain more than one POS tag and one of the POS tags is selected with respect to the chunk type. In the following chunk, the POS tag (n.animals) is selected with respect to the chunk type (NC).

For example:

NC [ခွေး/n.animals,တစ်/part.number,ကောင်/part.type]

If the noun chunk (NC) contains more than one noun, the last noun (n.food) is selected as a main word according to the nature of Myanmar language.

For example:

NC [ဆောင်းရာသီ/n.time,သီးနှံပင်/n.food,များ/part.number]

There are many possible function tags ($t_1$, $t_2 \ldots t_k$) for each POS tag with category (pc). These possible tags are retrieved from the training corpus by using the following equation that is prior probability as shown in figure 3.

$$\mathbf{P\ (t_k|pc) = C\ (t_k,pc)/C(pc)} \qquad \mathbf{(1)}$$

ppm.use#UseP:1.0

n.natural#PSubj:0.209,Subj:0.2985,PPla:0.1343,PObj:0.1642,PcomplS:0.0448,PPcomplO:0.0149,PCau:0.0448,PSim:0.0149,PAim:0.0299,Obj:0.0299,PCom:0.0149

pron.possessive#PIobj:0.1111,PSubj:0.2222,PObj:0.6667

cc.chunk#CCC:1.0

adj.dem#Ada: 0.9149, PObj: 0.0213,PSubj:0.0426,Active:0.0213

ppm.cause#CauP:1.0

n.verb#PSubj:0.6667,PObj:0.3333

v.common#Active:0.9744,VC:0.0128,PcomplS:0.0128

part.eg#PcomplOP:0.5455,SimP:0.4545

Figure 3. Sample data for POS/Function tag pairs with probability

We calculate the probability between next function tags ($n_1$, $n_2 \ldots n_j$) and previous possible tags by using the following equation that is log likelihood as shown in figure 4.

$$\mathbf{P\ (n_j|t_k) = C\ (n_j,t_k)/C(t_k)} \qquad \mathbf{(2)}$$

| |
|---|
| CCC,PSubj=0.2 |
| CCC,PAim=0.04 |
| CCC,Tim=0.04 |
| CCC,PcomplS=0.04 |
| PCau,CauP=1.0 |
| PPla,CCC=0.0156 |
| CCS,Ada=0.0196 |
| CCS,PTim=0.0196 |
| CCS,Tim=0.0131 |
| PlaP,Active=0.6111 |
| PlaP,Subj=0.1111 |

Figure 4. Sample data for Function/Function tag pairs with probability

Possible function tags are disambiguated by using Naïve Bayesian method. We multiply the probabilities from (1) and (2) and choose the function tag with the largest number as the posterior probability.

Technically, the task of function tags assignment is to generate a sentence that has correct function tags attached to certain words.

Our description of the function tagging process refers to the example as shown in figure 5, which illustrates the sentence ("မမနှင့်လှလှသည် ကျောင်းသို့ စက်ဘီးဖြင့် သွားသည်။" (Ma Ma and Hla Hla go to school by bicycle). This sentence is represented as a sequence of word-tags as "noun verb conjunction noun ppm pronoun verb". It is described as a sequence of chunk as "NC VC CC NC PPC NC VC SFC".

| |
|---|
| (a) NC[မမ/n.person]#CC[နှင့်/cc.chunk]#NC[လှလှ/n.person]#PPC[သည်/ppm.subj]#NC[ကျောင်း/n.location]#PPC[သို့/ppm.place]#NC[စက်ဘီး/n.objects]#PPC[ဖြင့်/ppm.use]#VC[သွား/v.common]#SFC[သည်/sf]။ |
| (b) PSubj[မမ]#CCC[နှင့်]#PSubj[လှလှ]#SubjP[သည်]#PPla[ကျောင်း]#PlaP[သို့]#PUse[စက်ဘီး]#UseP[ဖြင့်]#Active[သွားသည်]။ |

Figure 5. An overview of function tagging of the sentence
(a)The input POS-tagged and chunk sentence (b) The output sentence with function tags

## 8.2. Grammatical Relations of Myanmar Sentence

The LANGUAGE defined by a CFG (context-free grammar) is the set of strings derivable from the start symbol S (for Sentence). The core of a CFG grammar is a set of production rules that replaces single variables with strings of variables and symbols. The grammar generates all strings that, starting with a special start variable, can be obtained by applying the production rules until no variables remain. A CFG is usually thought in two ways: a device for generating sentences, or a device if assigning a structure to a given sentence. We use CFG for grammatical relations of function tags.

A CFG is a 4-tuple <N,Σ,P,S> consisting of

- A set of non-terminal symbols N
- A set of terminal symbols Σ
- A set of productions P
  - A-> α
  - A is a non-terminal
  - α is a string of symbols from the infinite set of strings (ΣU N)*
- A designated start symbol S

### 8.2.1. Simple Sentence

Consider a simple declarative sentence "သူတို့သည် မောင်ဘကို ခေါင်းဆောင် အဖြစ် ရွေးချယ်ခဲ့ သည်။" (They selected Mg Ba as a leader).

(a) NC[သူတို့/pron.possessive]#PPC[သည်/ppm.subj]#NC[မောင်ဘ/n.person]#PPC[ကို/ppm.obj]#NC [ခေါင်းဆောင်/n.person]#PPC[အဖြစ်/part.eg]#VC[ရွေးချယ်/v.common,ခဲ့/part.support]#SFC[သည်/ sf]။

(b) PSubj[သူတို့]#SubjP[သည်]#PObj[မောင်ဘ]#ObjP[ကို]#PPcomplO[ခေါင်းဆောင် ]#PcomplOP[အဖြစ်] # Active[ရွေးချယ်ခဲ့သည်]။

(c)

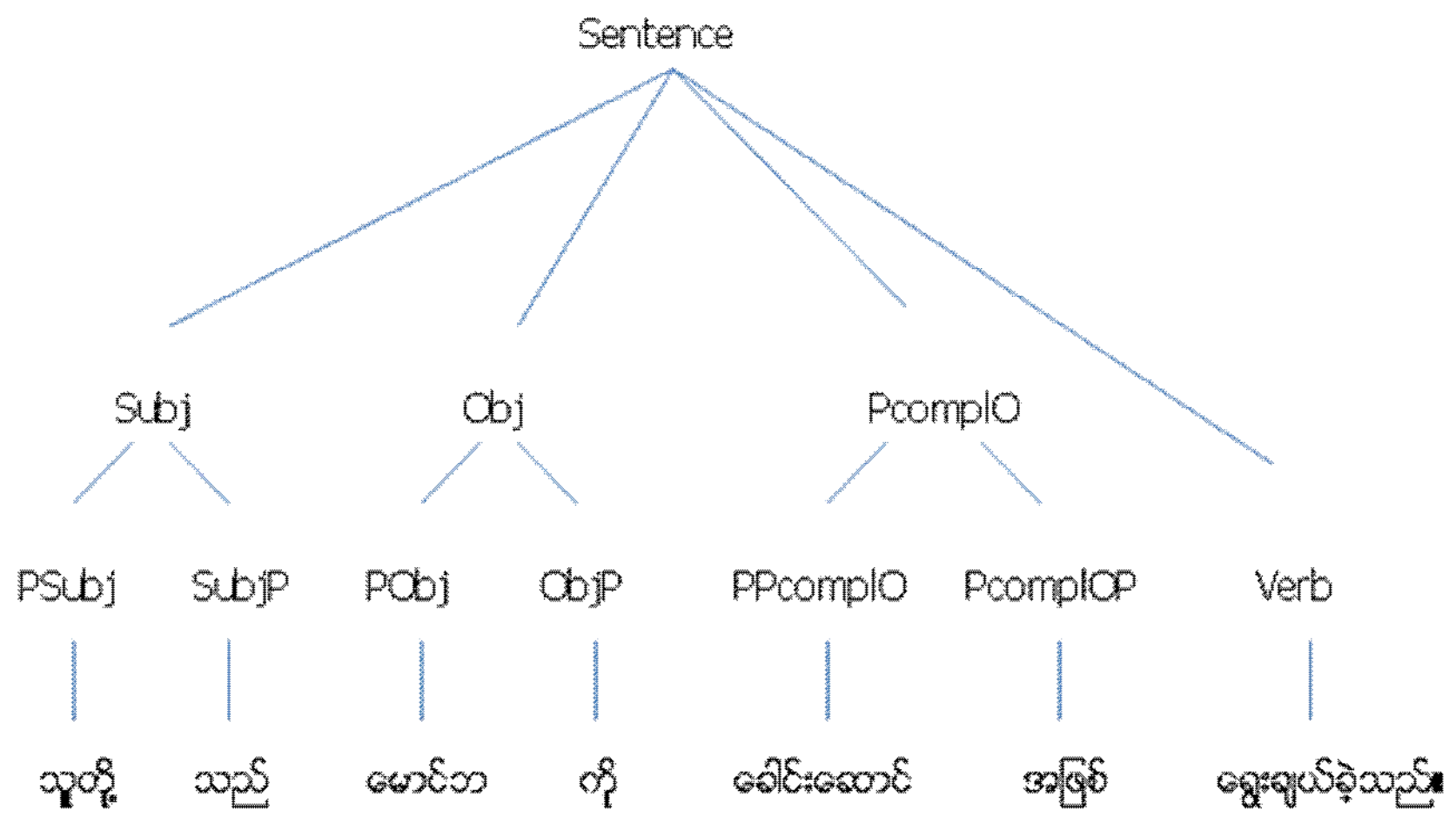

Figure 6. An overview of the function tagging and grammatical relations of simple sentence (a) The tagged and chunk sentence (b) The sentence with function tags (c) The syntactic tree structure with function tags

### 8.2.2. Complex Sentence

Our description of the parsing process refers to the example in figure 7, which illustrates the sentence "အဖေပေးသောစာအုပ်ကိုကျွန် တော်ဖတ်သည်။" (I read the book which is given by my father). This sentence is represented as a sequence of word-tags as "N V CC N PPC PRON V" .It is described as a sequence of chunk as "NC VC CC NC PPC NC VC SFC" and the sentence structure (Sentence) contains separate constituents for the object sentence (Obj-sent) and independent sentence (I-sent), which contains other phrases. Note that this parse tree has had some constituents conflated to comply with the constraint that there be only one constituent per word.

(a) NC [အဖေ/n.person] # VC [ပေး/v.common] # CC [သော/cc.adj] # NC [စာအုပ်/n.objects] # PPC [ကို/ppm.obj] # NC [ကျွန်တော်/pron.person] # VC [ဖတ်/v.common] # SFC [သည်/sf]။

(b) Subj[အဖေ]#Active[ပေး]#CCA[သော]#PObj[စာအုပ်]#ObjP[ကို]#Subj[ကျွန်တော်]#Active[ဖတ်သည်]။

(c)

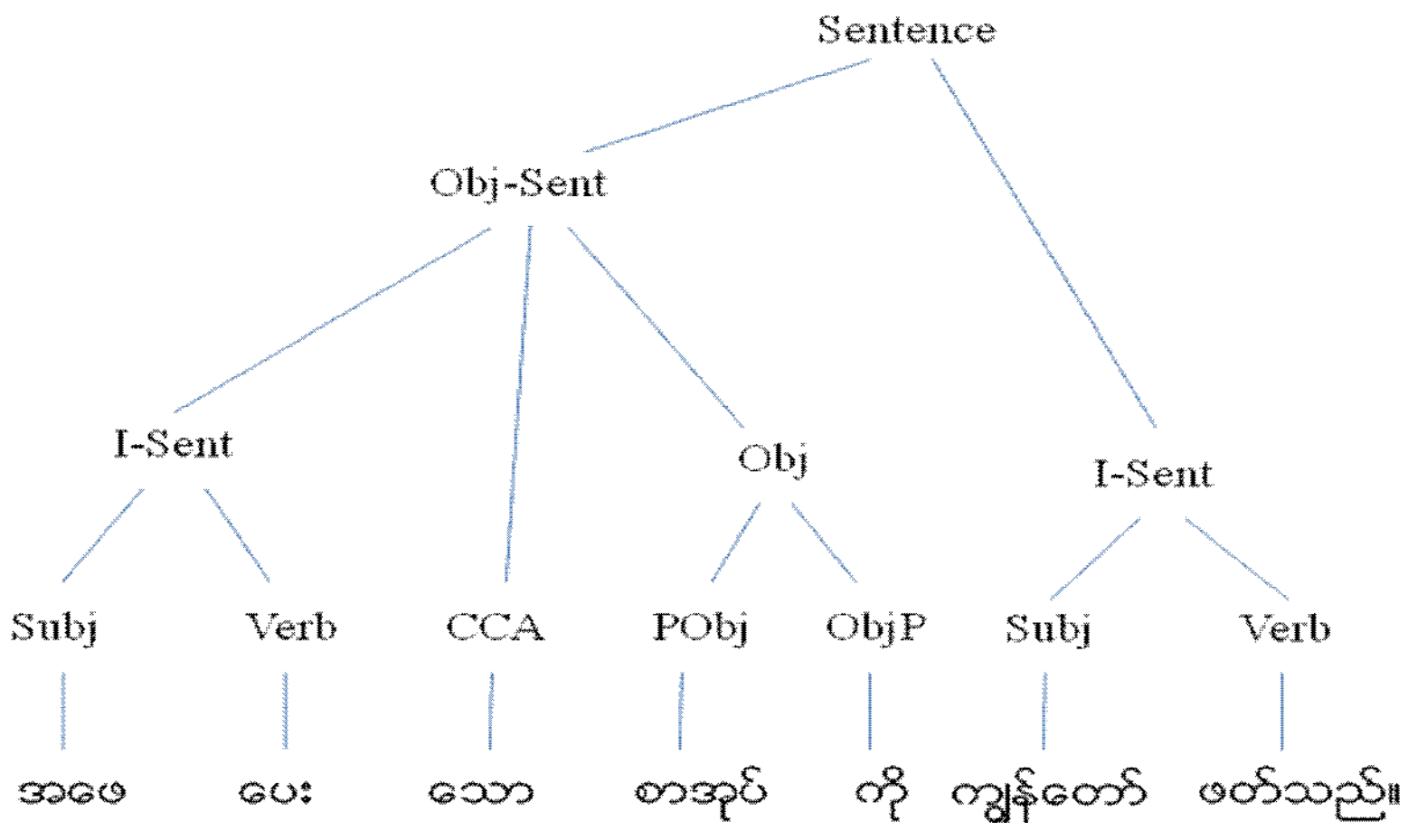

Figure 7. An overview of the function tagging and grammatical relations of complex sentence
(a) The tagged and chunk sentence (b) The sentence with function tags
(c) The syntactic tree structure with function tags

## 9. Evaluation and Results

The corpus contains about 3000 sentences with average word length 15. All sentences can be further classified as two sets. One is simple sentence set, in which every sentence has no more than 15 words. The other is complex sentence set, in which every sentence has more than 15 words. There are 1800 simple sentences and 1200 complex sentences in the corpus.

For evaluation purpose, different numbers of sentences collected from Myanmar textbooks of middle school and Myanmar grammar books are used as a test set. The test set can be divided into two groups: first group sentences are composed of word patterns in corpus and second group sentences are composed of word patterns that are not in the corpus. There are 60 sentences in the first group and 40 in the second one.The sentences are tested in the program and the function tagged results are manually checked. In table 8, the performance of function tagging according to the two groups is described.

Table 8. Performance of function tagging for different sentence patterns

| Sentence Patterns | Accuracy |
|---|---|
| sentence patterns in the corpus | 97.4% |
| sentence patterns that are not in the corpus | 89.6% |

After implementation of the system using the grammar, it has been seen that the system can easily generates the parse tree for a sentence if the sentence structure satisfies the grammar rules.

For example we take the following Myanmar simple sentence
မလှ သည် သူ့အမေ အတွက် ကိတ်မုန့် ဝယ်လာသည်။
(Ma Hla buys a cake for her mother.)

The structure of the above sentence is Subj-Aim-Obj-Pla-Verb. This is a correct sentence according to the Myanmar literature. According to the grammar a possible top-down derivation for the above simple sentence is

| | |
|---|---|
| 1. Sentence | [start] |
| 2. >>I-sent | [Sentence→I-sent ] |
| 3. >> Subj-Aim-Obj- Verb | [I-sent→Subj-Aim-Obj-Verb] |
| 4. >> PSubj SubjP -Aim-Obj- Verb | [Subj → PSubj SubjP] |
| 5. >> PSubj SubjP –PAim-AimP-Obj- Verb | [Aim → PAim AimP] |
| 6. >> PSubj SubjP –PAim-AimP-Obj-Verb | [Obj→Obj] |

For example we take the following Myanmar complex sentence
မိုးရွာလျှင်ကလေးများသည်လမ်းပေါ်တွင်ဘောလုံးကန်ကြသည်။
(If it rains, the children play the football on the road.)

The structure of the above sentence is Verb-CCS-Subj-Pla-Obj-Verb. This is a correct sentence according to the Myanmar literature. According to the grammar a possible top-down derivation for the above complex sentence is

| | |
|---|---|
| 1. Sentence | [start] |
| 2. >>I-sent CCS I-sent | [Sentence→I-sent CCS I-sent] |
| 3. >>Verb CCS I-sent | [I-sent→Verb] |
| 4. >>Verb CCS Subj Pla Obj Verb | [I-sent→Subj Pla Obj Verb] |
| 5. >>Verb CCS PSubj SubjP Pla Obj Verb | [Subj → PSubj SubjP] |
| 6. >>Verb CCS PSubj SubjP PPla PlaP Obj Verb | [Pla → PPla PlaP] |
| 7. >>Verb CCS PSubj SubjP PPla PlaP Obj Verb | [Obj→Obj] |

From the above derivation it has been seen that the Myanmar sentence is correct according to the grammar. So our system generates a parse tree successfully.

Our program tests only the sentence structure according to the grammar rules. So if the sentence structure satisfies the grammar rule, program recognizes the sentence as a correct sentence and generates a parse tree. Otherwise it gives output as an error.

## 10. CONCLUSION AND FUTURE WORK

In this paper, we investigate the function tag of the word depending on the sentence structure of Myanmar language. We used Naïve Bayesian technique for the task of assigning function tags. For grammatical relations of the function tags, we use context free grammar. The parse tree can be built by using function tags.

As function tagging is a pre-processing step for grammatical relations, the errors occurred in the task of function tagging affect the relations of the words. The corpus may be balanced because Naïve Bayesian framework probability simply describes uncertainty. The corpus creation is time consuming. The corpus is the resource for the development of Myanmar to English translation system and we expect the corpus to be continually expanded in the future because the tested sentence can be added into the corpus.

In this work we have considered limited number of Myanmar sentences to construct the grammar rules. In future work we have to consider as many sentences as we can and some more tags for constructing the grammar rules because Myanmar language is a free-phrase-order language. Word position for one sentence may not be same in the other sentences. So we can not restrict the grammar rules for some limited number of sentences.

**Authors**

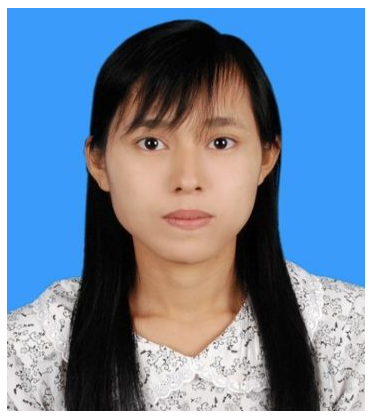

**Win Win Thant is** a Ph.D research student. She received B.C.Sc (Bachelor of Computer Science) degree in 2004, B.C.Sc (Hons.) degree in 2005 and M.C.Sc (Master of Computer Science) degree in 2007. She is now Assistant Lecturer of U.C.S.Y (University of Computer Studies, Yangon). She has written one local paper for Parallel and Soft Computing (PSC) conference in 2010 and one international paper for ICCA conference in 2011. Her research interests include Natural Language Processing and Machine Translation.

**Tin Myat Htwe is** an Associate Professor of U.C.S.Y. She obtained Ph.D degree of Information Technology from University of Computer Studies, Yangon. Her research interests include Natural Language Processing, Data Mining and Artificial Intelligence. She has published few papers in International conferences and International Journals.

**Ni Lar Thein** is a Rector of U.C.S.Y. She obtained B.Sc.(Chem.),B.Sc.(Hons) and M.Sc. (Computer Science) from Yangon University and Ph.D.(Computer Engg.) from Nanyang Technological University, Singapore in 2003. Her research interests include Software Engineering, Artificial Intelligence and Natural Language Processing. She has published few papers in International conferences and International Journals.